\documentclass[conference]{IEEEtran}
\IEEEoverridecommandlockouts
\usepackage{cite}
\usepackage{amsmath,amssymb,amsfonts}
\usepackage{algorithm,algpseudocode}
\usepackage{graphicx}
\usepackage{textcomp}
\usepackage{xcolor}
\usepackage[a4paper, total={184mm,239mm}]{geometry}
\usepackage{subcaption}
\usepackage{algorithm}
\usepackage{multirow}
\def\BibTeX{{\rm B\kern-.05em{\sc i\kern-.025em b}\kern-.08em
    T\kern-.1667em\lower.7ex\hbox{E}\kern-.125emX}}
\usepackage{adjustbox}
\usepackage{booktabs}
\usepackage{tikz}
\usepackage{xcolor}
\usepackage{hyperref}
\newcommand*\circled[1]{\tikz[baseline=(char.base)]{
            \node[shape=circle,fill,inner sep=0.8pt] (char) {\textcolor{white}{#1}};}}
\begin{document}

\renewcommand{\IEEEbibitemsep}{0pt plus 0.5pt}
\makeatletter
\IEEEtriggercmd{\reset@font\normalfont\fontsize{7pt}{7pt}\selectfont}
\makeatother
\IEEEtriggeratref{1}

\title{TT-SNN: Tensor Train Decomposition for Efficient Spiking Neural Network Training \\
\thanks{}
}

\author{Donghyun Lee, Ruokai Yin, Youngeun Kim, Abhishek Moitra, Yuhang Li, Priyadarshini Panda \\
Department of Electrical Engineering, Yale University, USA\\
\{donghyun.lee, ruokai.yin, youngeun.kim, abhishek.moitra, yuhang.li, priya.panda\}@yale.edu}



\maketitle

\begin{abstract}
 Spiking Neural Networks (SNNs) have gained significant attention as a potentially energy-efficient alternative for standard neural networks with their sparse binary activation. However, SNNs suffer from memory and computation overhead due to spatio-temporal dynamics and multiple backpropagation computations across timesteps during training. To address this issue, we introduce Tensor Train Decomposition for Spiking Neural Networks (TT-SNN), a method that reduces model size through trainable weight decomposition, resulting in reduced storage, FLOPs, and latency. In addition, we propose a parallel computation pipeline as an alternative to the typical sequential tensor computation, which can be flexibly integrated into various existing SNN architectures. To the best of our knowledge, this is the first of its kind application of tensor decomposition in SNNs. We validate our method using both static and dynamic datasets, CIFAR10/100 and N-Caltech101, respectively. We also propose a TT-SNN-tailored training accelerator to fully harness the parallelism in TT-SNN. Our results demonstrate substantial reductions in parameter size (7.98$\times$), FLOPs (9.25$\times$), training time (17.7\%), and training energy (28.3\%) during training for the N-Caltech101 dataset, with negligible accuracy degradation. Code is available at \href{https://github.com/donghyunli/TT-SNN}{Github}
\end{abstract}


\section{Introduction} \label{intro}

Spiking Neural Networks (SNNs) have gained significant interest as a low-power substitute to Artificial Neural Networks (ANNs) in the past decade \cite{roy2019towards}. Unlike ANNs, SNNs process visual data in an event-driven manner, employing sparse binary spikes across multiple timesteps.
This unique spike-driven processing mechanism brings high energy efficiency on various computing platforms \cite{davies2018loihi,sata}.
To leverage the energy-efficiency advantages of SNNs, many SNN training algorithms have been proposed, which can be categorized into two approaches: ANN-to-SNN conversion~\cite{diehl2015fast, han2020deep} and backpropagation (BP) with surrogate gradient~\cite{wu2018spatio, shrestha2018slayer}. Among them, BP-based training stands out as a mainstream training method as it not only achieves state-of-the-art performance but also requires a small number of timesteps ($\le5$). However, as BP-based training computes backward gradients across multiple timesteps and layers, SNNs require substantial training memory to store the intermediate activations~\cite{yin2023workload}.


\begin{figure*}
    \centering
    \includegraphics[width=0.8\linewidth]{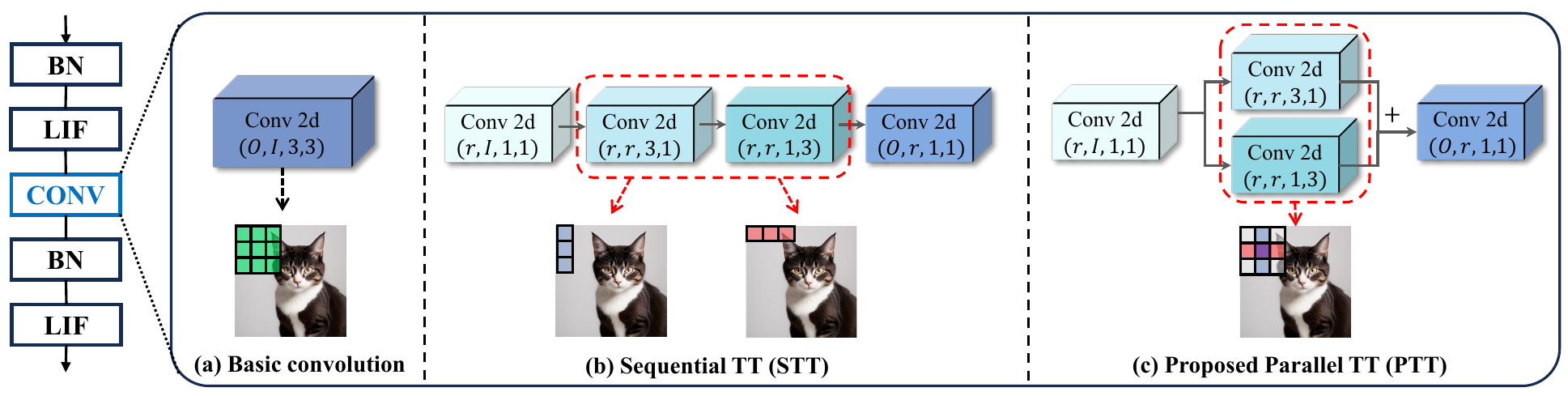}
    \caption{Illustration of TT-SNN modules. The order of weight information is followed according to the pytorch framework, i.e., (output channel, input channel, kernel size, kernel size) (a) Basic convolution weights with 3$\times$3 kernel. (b) Sequential computation of TT-cores is considered a traditional method with asymmetric kernels. (c) Proposed Parallel TT-module (PTT). Two asymmetric kernels are computed in parallel with the output of the first sub-convolution. The parallel computation of PTT can be seen as 3$\times$3 without the four corner values.}
    \label{fig1}
    \vspace{-5mm}
\end{figure*}

To address the challenge, various efficient techniques have been applied to SNNs, including quantization~\cite{putra2021q, yin2023mint}, Knowledge Distillation (KD)~\cite{xu2023constructing, kushawaha2021distilling}, and pruning~\cite{ yin2023workload}. In~\cite{putra2021q}, the authors focus on quantizing trainable weights to enable efficient and faster inference on SNN architecture. Additionally, prior research~\cite{yin2023mint} has explored the weight and membrane potential quantization to 2-bit for efficient hardware implementation. In terms of KD, in~\cite{xu2023constructing}, the ANN-based teacher model transfers the knowledge to the student model which is an SNN architecture for faster convergence. In contrast, in~\cite{kushawaha2021distilling}, the teacher model is SNN architecture, whose spike distribution is transferred to stabilize the training process of the student model. These efforts have proven successful in reducing memory costs and the total number of timesteps required, all while achieving specific accuracy targets. Nonetheless, the previous techniques mostly aim at fast and light inference, rather than focusing on training efficiency.



In this work, we introduce the TT-SNN module for accelerating training in SNNs by applying Tensor Train (TT) decomposition~\cite{oseledets2011tensor}. Our approach involves decomposing the weights of convolutional layers into smaller tensors, resulting in a lighter and faster training process. Inspired by~\cite{ding2019acnet}, we modify the computation pipeline with parallel asymmetric-sized kernels, which we term as Parallel TT (PTT), in contrast to the conventional Sequential TT (STT) operations. Additionally, we introduce the Half TT (HTT) module, which employs partial parts of PTT computations to boost training efficiency. Following the training process, we reconstruct the decomposed convolutional weights to maintain spike-inspired computation in the inference pipeline. 

From the hardware perspective, the recent advancements in SNN-tailored training accelerators, such as~\cite{sata} and~\cite{liang2021h2learn}, have paved the way for efficient SNN training. Given that the STT method operates on a single sub-convolutional layer at a time, it aligns well with the design of existing SNN training accelerators~\cite{sata,liang2021h2learn}, where each sub-convolutional layer is mapped sequentially onto the computation engines of the accelerator. This ensures a direct and efficient mapping without the need for significant architectural modifications. However, the PTT and HTT methods introduce a level of parallelism that is not optimally handled by the existing SNN training accelerators. The reason is that prior accelerators are primarily designed for single-layer workloads. The parallelism inherent in PTT and HTT, where two sub-convolutional layers operate concurrently, demands a more intricate hardware design. In this work, we propose a TT-SNN-tailored training accelerator design to fully harness the parallelism from the proposed PTT and HTT methods.


The main contributions of our work are as follows: (1) We propose the TT-SNN module, which leverages TT decomposition to enhance the efficiency of SNN architecture by enabling faster computation and reducing memory costs during training. Departing from the typical sequential computations, we introduce parallel processing into the training pipeline. (2) The TT-SNN module can be easily and flexibly integrated into SNN convolutional computations. (3) We propose a multi-cluster systolic-array-based SNN training accelerator to efficiently implement and evaluate the TT-SNN-based training. Compared to the existing SNN training accelerator, our design further reduces 28.3\%(43.5\%) energy cost for PTT(HTT) training. (4) Our experiments demonstrate that TT-SNN reduces the number of trainable parameters, floating-point operations (FLOPs), and training time on datasets such as CIFAR10/100 and N-Caltech101~\cite{orchard2015converting} without significant accuracy loss. For example, on the N-Caltech101 dataset, TT-SNN achieves compression ratios of 7.98$\times$ in parameters, 9.25$\times$ in FLOPs, and 17.66\% training time reduction, which highlights its compatibility with event datasets.

\section{Preliminary}

\noindent \textbf{Spiking Neural Network: }SNNs are brain-inspired architectures designed for efficient computation on neuromorphic devices. It relies on the Leaky-Integrate-and-Fire (LIF) neuron~\cite{izhikevich2003simple}, a non-linear function that closely mimics biological neurons in humans, making it an ideal choice for SNN design due to its efficient computation. We use the iterative LIF neuron model in~\cite{wu2018spatio} for designing SNN architecture as follows:
\vspace{-2mm}
\begin{equation}
    {u^{l,t}_i = \tau_m u^{l,t-1}_i+\sum^n_{j=1}w_{ij}H(u^{l,t}_i-V_{th})},
    \vspace{-2mm}
\label{lif1}
\end{equation}
where, ${u^{l,t}_i}$ is the membrane potential of $i$-th neuron in $l$-th layer at timestep $t$, $\tau_m\in(0,1]$ is the leaky factor for membrane potential leakage, $H(\cdot)$ is the Heaviside step function with firing threshold $V_{th}$. When a spike fires, $u^{l,t}_i\geq V_{th}$, the membrane potential ${u^{l,t}_i}$ is reset to 0. 




\noindent \textbf{Tensor Train Decomposition: }Tensor decomposition has emerged as a promising compression technique for mitigating the high redundancy inherent in Deep Neural Networks (DNNs) by reducing the number of trainable parameters. There are primarily three distinct types of tensor decomposition techniques applied to DNNs~\cite{liu2023tensor}, i.e., CANDECOMP/PARAFAC (CP), Tucker, and TT decomposition. While CP and Tucker decompositions are compact and efficient, they may not be suitable for large-scale models due to the curse of dimensionality~\cite{cichocki2015tensor}. In contrast, TT decomposition is less sensitive to the curse of dimensionality and remains a powerful tool for reducing the number of trainable parameters, that can be expressed as
\begin{equation}
    \mathcal{A} = \mathcal{G}_1 \times ^{1} \mathcal{G}_2 \times ^{1} \cdots \times ^{1} \mathcal{G}_d,
\label{TT-format}
\end{equation}
where $\mathcal{A} \in \mathbb{R}^{n_1 \times n_2 \times \cdots \times n_d}$ is a $d$-dimensional target tensor, $\mathcal{G}_k\in \mathbb{R}^{r_{k-1} \times n_k \times r_k}$ ($k \in \{1,2, \cdots ,d\}$) is the $k$-th decomposed tensor, and $\times ^{1}$ denotes the contraction. $\mathcal{G}_k$ is called as TT-core and  $r_k(r_0=r_d=1)$ is TT-rank which controls the complexity of decomposition. 
While TT decomposition has been applied to several DNN structures~\cite{wang2020compressing, lee2021qttnet}, most of these applications mainly concentrate on reducing theoretical computation complexity rather than addressing actual hardware latency. In contrast, Gabor et al.~\cite{gabor2022convolutional} successfully reduce FLOPs in each CNN layer by incorporating circular permute into the TT decomposition.
\begin{equation}
    \mathcal{W} = \texttt{circular\_permute}(W, -1) \in \mathbb{R}^{I \times K \times K\times O}
\end{equation}
Here, $W$ refers to CNN weights, $I$ is the number of input channels, $O$ is the number of output channels, and $K$ is kernel size. Then, $\mathcal{W}$ can be decomposed by TT-format according to Eq.\eqref{TT-format}, which can be represented as
\begin{equation}
    \mathcal{W}_{I,K_1,K_2,O}=\sum^{R_1}_{r_1}\sum^{R_2}_{r_2}\sum^{R_3}_{r_3} w^{(1)}_{I,r_1}w^{(2)}_{r_1,K_1,r_2}w^{(3)}_{r_2,K_2,r_3}w^{(4)}_{r_3,O}.
\label{decompose}
\end{equation}
Based on~\cite{gabor2022convolutional}, one convolution computation can be separated into four sub-convolutions with smaller weights like Fig.~\ref{fig1}(b), which results in a faster and lighter training process. This approach is different from the TT decomposition used in previous works~\cite{wang2020compressing, lee2021qttnet}, where an extra reconstruction process of TT-cores is applied in every convolutional operation during both training and inference.


\section{Proposed Method}
In this section, we begin by introducing the TT-SNN framework, which incorporates TT decomposition into SNNs. The first TT module is the PTT, designed to parallelize convolutional computations. Additionally, we present another TT module, HTT, specifically engineered to enable half-diagonal computation within the spatio-temporal computation dimension. Finally, we demonstrate the overall training process of TT-SNN architecture. 


\noindent \textbf{Parallel TT Module: }To address the challenges posed by the high training complexity of SNNs, as discussed in Section \ref{intro}, we introduce a novel and straightforward training pipeline for SNNs called TT-SNN, illustrated in Fig. \ref{fig1}. TT decomposition allows us to break down one convolution into four sub-convolutions. In~\cite{gabor2022convolutional}, they sequentially compute the sub-convolutions, like Fig. \ref{fig1}(b), resulting in considerable performance with a reduction in the number of parameters and FLOPs. However, the Sequential TT (STT) loses the input information due to its asymmetric kernel size. The second and third sub-convolutions in STT compute with $(3,1)$ and $(1,3)$ kernels, leading to either vertical or horizontal information extraction. Consequently, the perpendicular direction information of each kernel from the previous layer is overlooked. 


In order to address this asymmetry in STT, we propose a simple modification, called Parallel TT (PTT) illustrated in Fig. \ref{fig1}(c). In the PTT pipeline, the second and third sub-convolution layers are computed in parallel, both utilizing the output from the first sub-convolution layer. This parallel computation in PTT resembles a 3$\times$3 kernel without the four corner values. Hence, this cross-sectional kernel can perceive the vertical and horizontal feature information simultaneously, resulting in improved performance. Motivated by~\cite{ding2019acnet}, we restructure the computations of sub-convolutions using Eq. \eqref{decompose} as follows:
\begin{equation}
    y_t = [(x_t * w^{(1)}_{I,r} * w^{(2)}_{r,K_1,r})+(x_t * w^{(1)}_{I,r}*w^{(3)}_{r,K_2,r})] *       w^{(4)}_{r,O},
\label{ATT}
\end{equation}
where $x_t, y_t$ are input and output in timestep $t$ respectively, and $*$ denotes convolution computation. 


\begin{figure}
    \centering
    \subfloat[Half TT (HTT)]{\includegraphics[width=6.5cm]{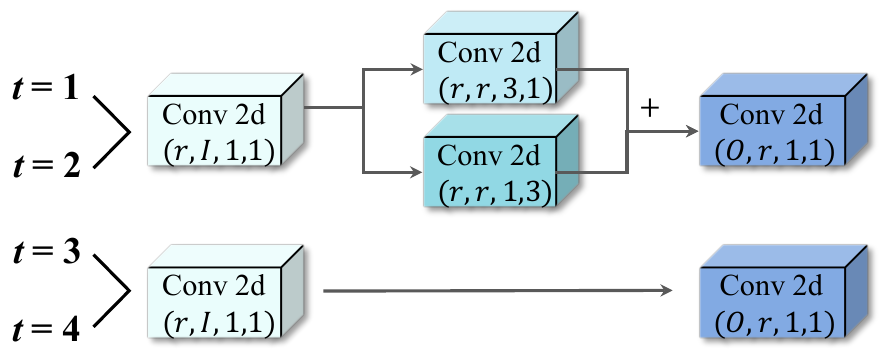}}
    \hspace{1mm}
    \subfloat[Spatio-temporal computation diagram]{\includegraphics[width=7cm]{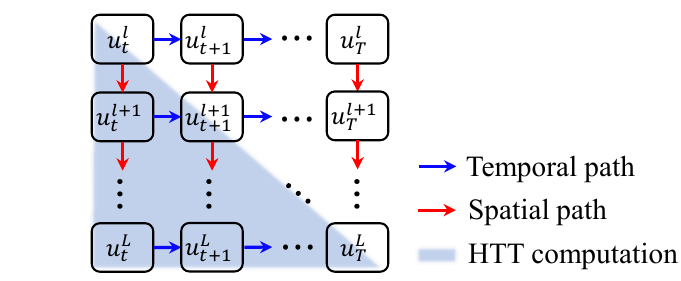}}
    \caption{Illustration of Half TT (HTT) format for further compression. (a) Instead of sharing all weights through timesteps, HTT uses partial parts of sub-convolutions. (b) In the spatio-temporal computation dimension of SNN, the HTT module takes up a half-diagonal area due to its partial usage of weights through timestep.}
    \label{fig2}
    \vspace{-5mm}
\end{figure}

\noindent \textbf{Half TT Module: }In traditional SNN architectures, weight sharing across timesteps is a common approach to keep weight storage consistent, even as the number of timesteps increases. \cite{kim2023exploring} finds that SNNs tend to capture more information during the early timesteps compared to the later ones, implying the existence of redundancy in timesteps. To explore this argument, we introduce a novel strategy called the Half TT (HTT) module, which operates by using only half of the sub-convolutions in select timesteps, as opposed to employing all sub-convolutions throughout all timesteps as depicted in Fig.~\ref{fig2}(a). The main difference between PTT and HTT is that PTT employs all sub-convolutions throughout the entire timestep, while HTT uses half of the sub-convolutions in specific timesteps. We place full sub-convolutions in the early timesteps and reserve half sub-convolutions for the later timesteps. In this approach, we can weaken the timestep redundancy and achieve faster and more resource-efficient computation. The HTT can be seen as a half-diagonal computation in a spatio-temporal computation graph diagram on SNN shown in Fig.~\ref{fig2}(b).



\noindent \textbf{Training Pipeline: }
By aggregating the proposed TT modules, we represent an efficient training pipeline, as outlined in Algorithm 1. To begin, we initialize the base SNN model to gain optimized TT-ranks using Variational Bayesian Matrix Factorization (VBMF)~\cite{nakajima2013global}. VBMF is a practical tool for estimating near-optimal ranks with automatic posterior approximation. Subsequently, the TT-SNN model is initialized with decomposed weights and the acquired TT-ranks (lines 1-5). Note that the first CNN layer and the last classifier are not decomposed layers as customization of these layers results in a significant drop in accuracy. After training (lines 6-18), the entire weights of sub-convolutions are merged into a single original weight to enable spike-based computations throughout the model (lines 19-21). The Eq. \eqref{linear} shows the reconstruction process from Eq. \eqref{ATT}. We simplify the weight terms in Eq. \eqref{ATT} from \{$w^{(1)}_{I,r_1}, w^{(2)}_{r_1,K_1,r_2}, w^{(3)}_{r_2,K_2,r_3}, w^{(4)}_{r_3,O}$\} to \{$w^{(1)}, w^{(2)}, w^{(3)}, w^{(4)}$\}.
\begin{equation}
\begin{split}
    y_t & = [(x_t * w^{(1)}* w^{(2)})+(x_t * w^{(1)}*w^{(3)}) ] * w^{(4)}\\    
        & = [x_t * (w^{(1)} \times^1 w^{(2)}) + x_t * (w^{(1)} \times^1 w^{(3)})]*w^{(4)} \\
        & = x_t*(w^{(1)} \times^1 w^{(2)}\times^1 w^{(4)}+w^{(1)} \times^1 w^{(3)}\times^1 w^{(4)}) \\
        & = x_t * \widetilde{\mathcal{W}}.
\end{split}
\label{linear}
\end{equation}
In summary, our proposed approach offers substantial advantages during training, and pre-trained TT modules can be converted into the base model architecture without incurring any significant losses.

\begin{algorithm}
\label{algo1}
\footnotesize
\caption{Training process of TT-SNN}

$l$, $t$, FC, LIF, and BN represent the $l$-th layer and the $t$-th timestep, the fully-connected layer, the LIF neuron model in Eq.~\eqref{lif1}, and the batch normalization respectively 
\begin{algorithmic}[1]     
    \State $[\mathcal{W}_1,\mathcal{W}_2, \ldots, \mathcal{W}_L]$ = Initialize(base model)
    \State $[r_2, r_3, \ldots, r_{L-1}]$ = VBMF($[\mathcal{W}_2,\mathcal{W}_3, \ldots, \mathcal{W}_{L-1}]$)
    \For{$l \leftarrow$ 2 to $L-1$} 
        \State $[w^{(1)}_l, w^{(2)}_l, w^{(3)}_l, w^{(4)}_l]$ = Initialize($\mathcal{W}_l$, $r_l$)
    \EndFor    
    \For{epochs} 
        \For{$t \leftarrow$ 1 to $T$}
            \State $y_{t,1} =$ $x_{t,1} * \mathcal{W}_1$
            \For{$l \leftarrow$ 2 to $L-1$}
                \State $x_{t,l} = $ BN(LIF($y_{t,l-1}$))
                \State $o_{t,l} = x_{t,l} * w^{(1)}_l$
                \State $y_{t,l} = [(o_{t,l} * w^{(2)}_l)+(o_{t,l} *w^{(3)}_l)] * w^{(4)}_l$
            \EndFor
            \State $y_{t,L} =$ FC(LIF($y_{t,L-1}$))
        \EndFor        
        \State $L=$ Cross-Entropy($\sum^T_{t=1}y_{t,L}$, $label$)
        \State Calculate $\frac{\partial L}{\partial w^{(1)}_l }, \frac{\partial L}{\partial w^{(2)}_l}, \frac{\partial L}{\partial w^{(3)}_l }, \frac{\partial L}{\partial w^{(4)}_l }$
        \State Update $w^{(1)}_l, w^{(2)}_l, w^{(3)}_l, w^{(4)}_l$
    \EndFor    
    \For{$l \leftarrow$ 2 to $L-1$}
        \State $\widetilde{\mathcal{W}_l} = (w^{(1)}_l \times^1 w^{(2)}_l\times^1 w^{(4)}_l) + (w^{(1)}_l \times^1 w^{(3)}_l\times^1 w^{(4)}_l)$
    \EndFor     
\end{algorithmic}
\end{algorithm}
\vspace{-2mm}

\section{SNN Training Accelerator Design for TT-SNN}
\label{sec:4-1}

\begin{figure}[t]
    \centering
    \includegraphics[width=\linewidth]{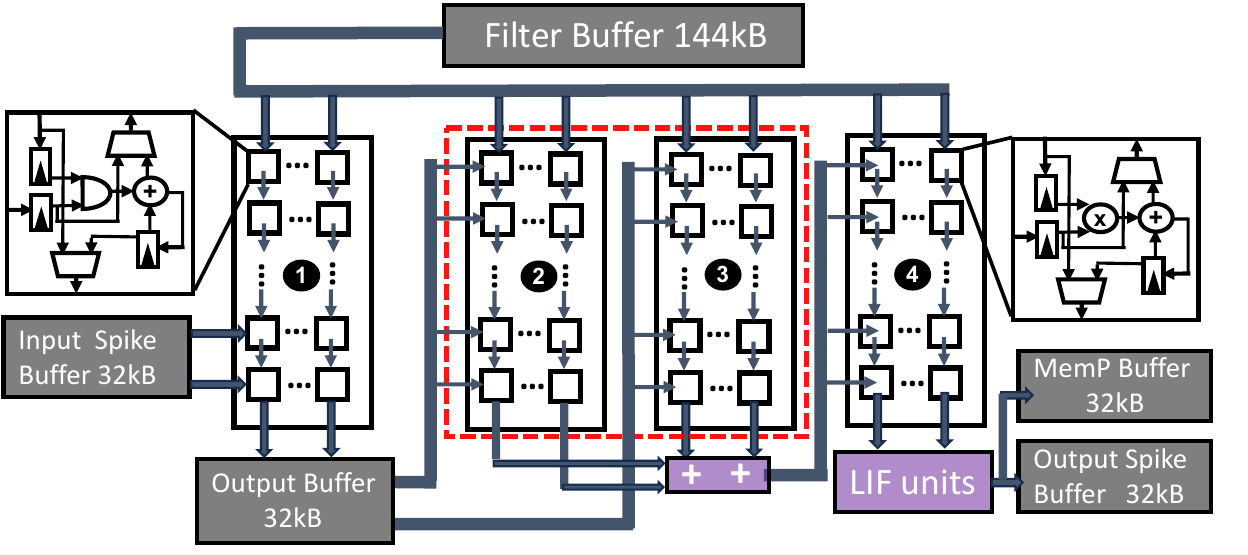}
    \caption{Illustration of the design of our training accelerator for efficiently mapping the PTT-SNN and HTT-SNN. MemP denotes the membrane potential.}
    \label{fig:arch}
    \vspace{-6.5mm}
\end{figure}

To fully harness the parallelization from the PTT and HTT methods, we propose a multi-cluster systolic-array-based SNN training accelerator design as shown in Fig.~\ref{fig:arch}. In our design, we have 4 computation clusters for mapping the workload of each sub-convolutional layer. The cluster~\circled{1} computes the first sub-convolutional layer. Since the input is in the form of spikes, we simplified the arithmetic units inside the PEs of~\circled{1}. As shown in Fig.~\ref{fig:arch}, cluster~\circled{2} and cluster~\circled{3} run in parallel (shown in the red rectangle). The outputs from \circled{1} will first be written into an output buffer and then consumed by the clusters~\circled{2} and \circled{3}. The generated results are then merged in the adder array units and sent to the cluster~\circled{4} for the computation of the last sub-convolutional layer. To support the non-spike inputs to those layers, we equip the PEs with multipliers in those three clusters~\circled{2} -~\circled{4}. The outputs from the last cluster will be sent to the LIF array units to be converted back to forms of spikes. Finally, both the spikes and the membrane potentials from the LIF units will be written back to the corresponding global buffers. We run the whole design in a highly pipelined fashion. While writing the inputs and weights to~\circled{1}, the weights are also being filled to the cluster~\circled{2} and \circled{3} to hide the SRAM read latency. The outputs from the global output buffer after the cluster~\circled{1} are instantly consumed by the parallel clusters \circled{2}, \circled{3}. The results from the adder arrays are also instantly consumed by the cluster~\circled{4}.

 We adopt three levels of the memory hierarchy: 1) an off-chip DRAM, 2) SRAM-based global buffers, and 3) small register-file-based scratch pads. The output-stationary dataflow is adopted in cluster~\circled{1} and \circled{4} and the weight-stationary dataflow is adopted in cluster~\circled{2} and \circled{3} for matching the latency between clusters. We will finish processing all timesteps for each layer and then move to the next~\cite{narayanan2020spinalflow}. We follow the dataflow in ~\cite{sata} and \cite{liang2021h2learn} to accelerate the BPTT-based backpropagation. The detailed design configurations of our design can be found in Table.~\ref{tab:parameter}.

 \begin{table}[h]
     \aboverulesep=0.1ex 
   \belowrulesep=0ex 
    \centering
    \caption{Hardware Implementation Parameters.}
    \begin{adjustbox}{max width =\linewidth}
	\begin{tabular}{lr}
        \toprule
        Technology &$28$nm CMOS\\
        \midrule
        \# of Cluster &4\\
        \midrule
        \# of PE / Cluster &32\\
        \midrule
        Scratch Pad Size / PE & 32 bytes\\
        \midrule
        Total Global Buffer Size & 272 KB\\
        \midrule
        Accumulator Precision &16-bits\\
        \midrule
        Multiplier Precision &8-bits\\
        \bottomrule
 	\end{tabular}\label{tab:parameter}
 \end{adjustbox}
 \vspace{-5mm}
\end{table}

\begin{table*}[t]
\caption{Results on CIFAR10/100 with ResNet18 and N-Caltech101 with ResNet34 architecture. All metrics are computed during the training process. Here, training time represents the time taken for forward and backward passes on a single batch of inputs, M and G denote millions and gigabytes respectively.}
\begin{center}
\begin{tabular}{c|c|c|c|c|c}
\hline
\textbf{Dataset}& \textbf{Method} & \textbf{Accuracy (\%)}  & \textbf{Training time (s)} & \textbf{\# of parameters (M)} & \textbf{FLOPs (G)} \\
\hline
& baseline & 93.41  & 0.214 & 11.20 & 2.221 \\
CIFAR10& STT & 90.91 & 0.190 (11.21 \% $\downarrow$) & 1.83 (6.13 $\times$) & 0.372 (5.97 $\times$) \\
(t = 4) & PTT & 91.65 & 0.176 (17.76 \% $\downarrow$) & 1.83 (6.13 $\times$) & 0.372 (5.97 $\times$) \\
& HTT & 91.19 & 0.166 (22.43 \% $\downarrow$) & 1.83 (6.13 $\times$) & 0.282 (7.88 $\times$) \\
\hline
& baseline & 72.49 & 0.214 & 11.21 & 2.222 \\
CIFAR100& STT & 68.49 & 0.190 (11.21 \% $\downarrow$) & 1.69 (6.62 $\times$) & 0.373 (5.96 $\times$) \\
(t = 4) & PTT & 70.44 & 0.176 (17.76 \% $\downarrow$) & 1.69 (6.62 $\times$) & 0.373 (5.96 $\times$) \\
& HTT & 70.22 & 0.167 (21.96 \% $\downarrow$) & 1.69 (6.62 $\times$) & 0.282 (7.87 $\times$) \\
\hline
& baseline & 77.13 & 0.657 & 21.31 & 15.65 \\
N-Caltech101& STT & 76.48 & 0.572 (12.94 \% $\downarrow$) & 2.67 (7.98 $\times$) & 1.69 (9.25 $\times$) \\
(t = 6) & PTT & 77.24 & 0.541 (17.66 \% $\downarrow$) & 2.67 (7.98 $\times$) & 1.69 (9.25 $\times$) \\
& HTT & 75.38 & 0.530 (19.33 \% $\downarrow$) & 2.67 (7.98 $\times$) & 1.46 (10.75 $\times$) \\
\hline
\end{tabular}
\label{tab1}
\end{center}
\vspace{-4.5mm}
\end{table*}

\begin{table}[t]
\caption{Training performance comparison before and after applying PTT to previous works with CIFAR10 and DVS Gesture datasets.}
\vspace{-3mm}
\begin{center}
\resizebox{0.485\textwidth}{!}{
\begin{tabular}{ccccc}
\hline
\multirow{2}{*}{\textbf{Method}} & \multirow{2}{*}{\textbf{Model}} & \multirow{2}{*}{\textbf{Dataset}}  & \multirow{2}{*}{\begin{tabular}[c]{@{}c@{}}\textbf{Accuracy(\%)}\\ (Base / PTT)\end{tabular}} & \multirow{2}{*}{\begin{tabular}[c]{@{}c@{}}\textbf{Training time(s)}\\ (Base / PTT)\end{tabular}} \\ \\
\hline
tdBN\cite{zheng2021going} & ResNet20 & CIFAR10 & 92.96 / 91.10 & 0.116 / 0.087 \\
TEBN\cite{duan2022temporal} & VGG9 & CIFAR10  & 91.78 / 90.56 & 0.066 / 0.056 \\
TET\cite{deng2022temporal} & VGG9 & DVS Gesture  & 94.79 / 94.49  & 0.351 / 0.319 \\
NDA\cite{li2022neuromorphic} & VGG11 & DVS Gesture & 96.88 / 95.83 & 0.299 / 0.240  \\
\hline
\end{tabular}
}
\label{tab2}
\end{center}
\vspace{-7mm}
\end{table}

\section{Experiments}
\subsection{Implementation Details}
\noindent \textbf{Software: }We evaluate our work on CIFAR10/100 with ResNet18 and N-Caltech101 with ResNet34 architecture, respectively. We have adopted MS-ResNet~\cite{hu2021advancing} as our baseline SNN architecture. We use direct coding to convert a float pixel value into binary spikes~\cite{wu2019direct}. During the training process, we use SGD optimizer with momentum 0.9 and weight decay 1e-4. We adopt cosine annealing scheduler for learning rate decay with initial learning rate 0.1. We set the number of epochs as 100 for all datasets. The batch size for CIFAR10, CIFAR100, and N-Caltech101 is set to 100, 100, and 50 respectively. In terms of spiking mechanism, we set $\tau_m$ and $V_{th}$ to 0.25, 0.5 respectively in Eq. \eqref{lif1}. The TT-ranks attained by VBMF for ResNet18 is $\{24,27,25,29,37,45,43,41,65,74,70,63,104,153,186,145\}$, and for ResNet34 is $\{24,23,22,17,16,12,22,31,25,25,24,21,\\20,19,48,79,64,69,63,69,60,65,63,63,62,58,121,170,173,\\147,161,108\}$. The training timestep is 4 for CIFAR10 and 6 for N-Caltech101. When we use the HTT module for training CIAFAR10/100 and N-Caltech101, half sub-convolutions are applied in timestep $t=3,4$ and $t=5,6$ respectively. All experiments are conducted by RTX 3090ti GPUs.


\noindent \textbf{Hardware: }We synthesize our accelerator in Sec.~\ref{sec:4-1} using Synopsys Design Compiler at 400MHz using 28nm CMOS technology. We use CACTI to simulate on-chip SRAM and off-chip DRAM to obtain memory statistics. The energy results are generated from SATASim, a cycle-accurate SNN training energy simulator~\cite{sata}. The training energy includes the computation and the data movement cost for both the forward and the backward propagation of one image across all timesteps.

\subsection{Experimental Results}
We have summarized the main results for the CIFAR10/100, N-Caltech101 datasets in Table \ref{tab1}, which includes accuracy, training time, the number of trainable parameters, FLOPs. Note that the training time denotes the time taken for forward and backward passes on a single batch of inputs of a given dataset.


\noindent \textbf{CIFAR10/100: }The results for CIFAR10/100 exhibit very similar trends across all evaluation metrics. As anticipated, among the TT modules, PTT achieves the highest accuracy but also an insignificant accuracy drop compared to the baseline. In terms of training time, PTT significantly reduces the training time, approximately 17\% faster than the baseline, and outperforms STT. All three modules have the same number of parameters, and the differences in training time stem from architectural modifications. For the CIFAR dataset, HTT emerges as the most efficient module, reducing training latency by over 21\% and FLOPs by about 8$\times$ compared to the baseline.

\noindent \textbf{N-Caltech101: }N-Caltech101 displays comparable trends with the CIFAR dataset across most evaluation metrics. However, the results for the HTT module differ. In contrast to the static CIFAR dataset, the accuracy of HTT is even lower than that of STT in N-Caltech101. We believe this accuracy drop is caused by the characteristics of dynamic datasets. When training an SNN architecture with static datasets, the input data remains consistent throughout the timesteps. Consequently, even with some half sub-convolutions in the later timesteps, the full sub-convolutions in the early timesteps suffice to extract feature information. However, in dynamic datasets, each input in every timestep is distinct. As a result, information loss occurs when half sub-convolutions are applied, as some information from the new input may not be effectively extracted. On the other hand, the accuracy of the PTT module surpasses that of the baseline and reduces FLOPs by about 9$\times$. This indicates that our proposed method integrates well with not only static datasets but also dynamic event datasets.

\begin{figure}[t]
\begin{center}
\def\arraystretch{0.5}
\begin{tabular}{@{}c@{}c}
\includegraphics[width=0.60\linewidth]{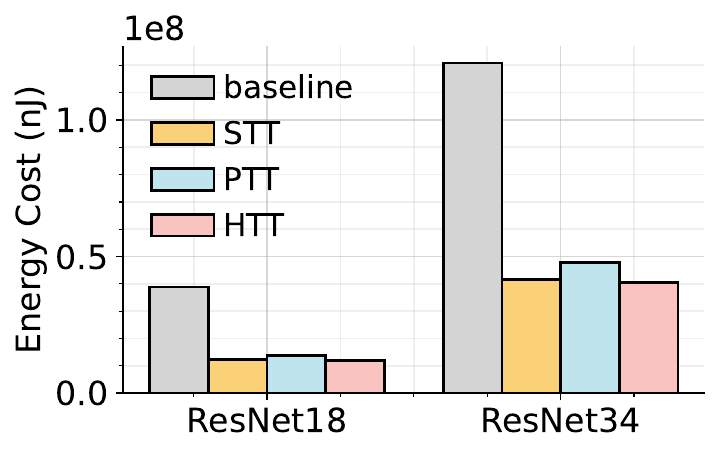} &
\includegraphics[width=0.35\linewidth]{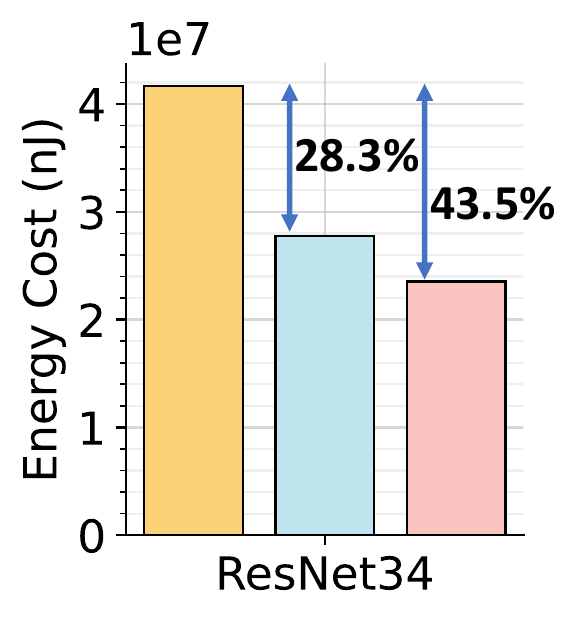}

\\
{ (a)} & { (b)}
\end{tabular}
\caption{(a) Training energy costs of STT, PTT, and HTT-based SNNs compared to the baseline SNN on ResNet18 and ResNet34. The results are calculated based on the accelerator design of~\cite{sata}. (b) The training energy cost improvements of PTT and HTT compared to STT on our proposed multi-cluster accelerator design.
}
\vspace{-8mm}
\label{fig:hw_results}
\end{center}
\end{figure}
\noindent \textbf{On the Existing SNN Training Accelerator: }We first directly simulate the energy costs on the existing SNN training accelerator~\cite{sata} for training the baseline SNNs, the STT-based SNNs, the PTT-based SNNs, and the HTT-based SNNs. The results are shown in Fig.~\ref{fig:hw_results}(a). Due to the model size reduction brought by the decomposition, we observe that the STT-based methods reduce $68.1\%$ training energy cost from the baseline SNNs. However, as we discussed in Sec.~\ref{sec:4-1}, due to the layer-by-layer mapping strategy in the prior works, the PTT-based and HTT-based SNNs do not benefit from the parallelism during the training. Consequently, compared to the training of STT-based SNNs, HTT-based SNNs cost similar energy, and the PTT-based SNNs even cost $10.9\%$ higher energy. The reason for the higher energy cost is the fact that the PTT method needs to store the outputs from one of the sub-convolutional layers to DRAM and then re-fetch them back to the chip to merge the results before proceeding to the last sub-convolutional layer.\\
\noindent \textbf{On the Proposed SNN Training Accelerator: } Since our proposed multi-cluster design can fully harness the parallelism between sub-convolutional layers, we manage to reduce $28.3\%$($43.5\%$) training energy cost on the PTT(HTT) method from the STT method as shown in Fig.~\ref{fig:hw_results}(b).

\vspace{-1mm}
\subsection{Compatibility with other SNN architectures}
We further verify the compatibility of our work with previous SNN architectures, including tdBN~\cite{zheng2021going}, TEBN~\cite{duan2022temporal}, TET~\cite{deng2022temporal}, and NDA~\cite{li2022neuromorphic}, by integrating the PTT modules, which is shown in Table ~\ref{tab2}. We utilize the architectures given by each previous work on both static and dynamic datasets: CIFAR10 and DVS128 Gesture~\cite{amir2017low}. The PTT module can decrease the training time in all methods: 25.00\% on tdBN, 15.15\% on TEBN, 9.12\% on TET, and 19.73\% on NDA respectively, without significant accuracy degradation. These results highlight the effectiveness and flexibility of TT-SNN as a powerful plug-in tool for accelerating the training process of any SNN algorithm.

\begin{figure}[t]
    \centering
    \subfloat{\includegraphics[width=4.25cm]{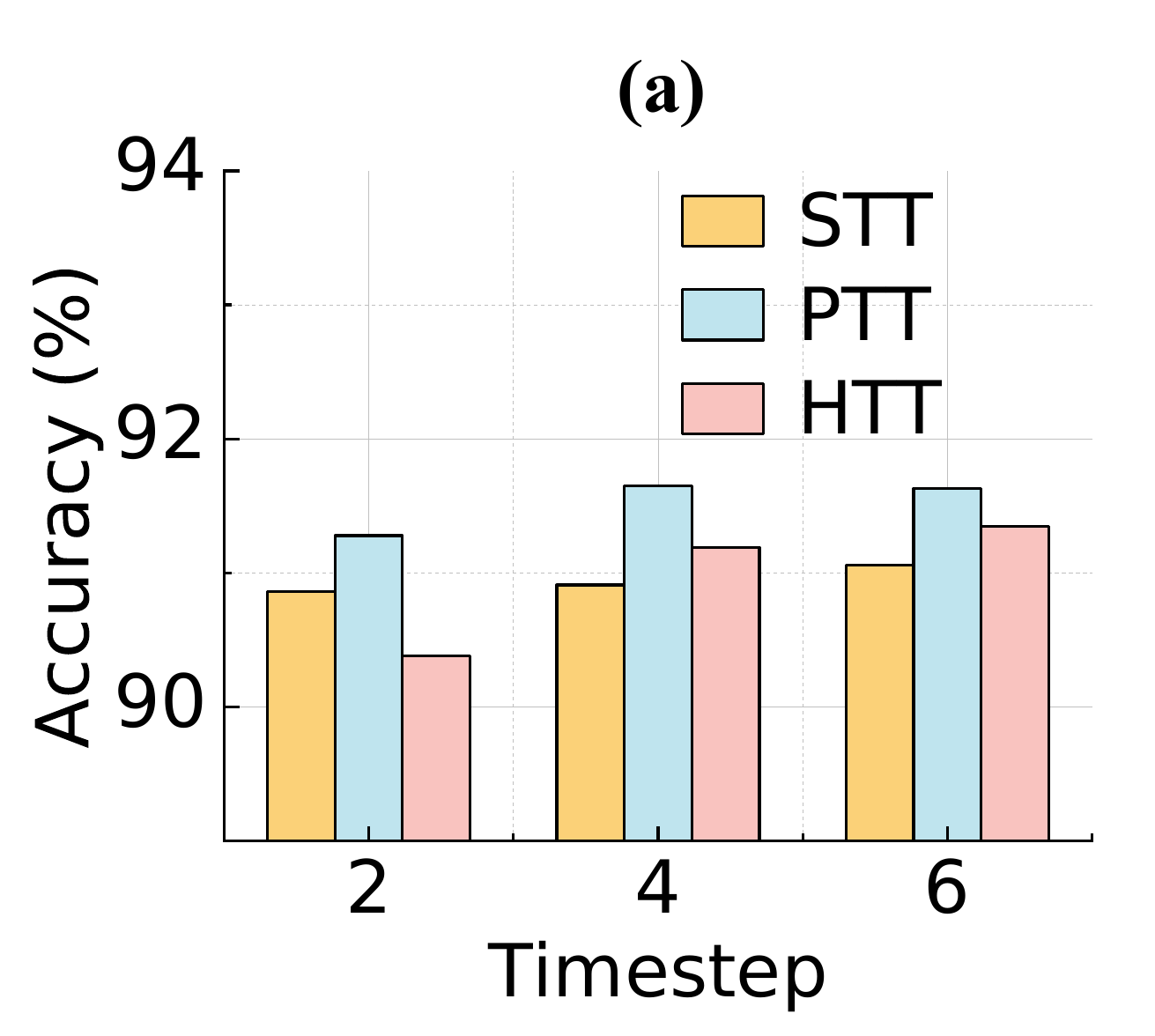}}
    \hspace{2mm}
    \subfloat{\includegraphics[width=4.0cm]{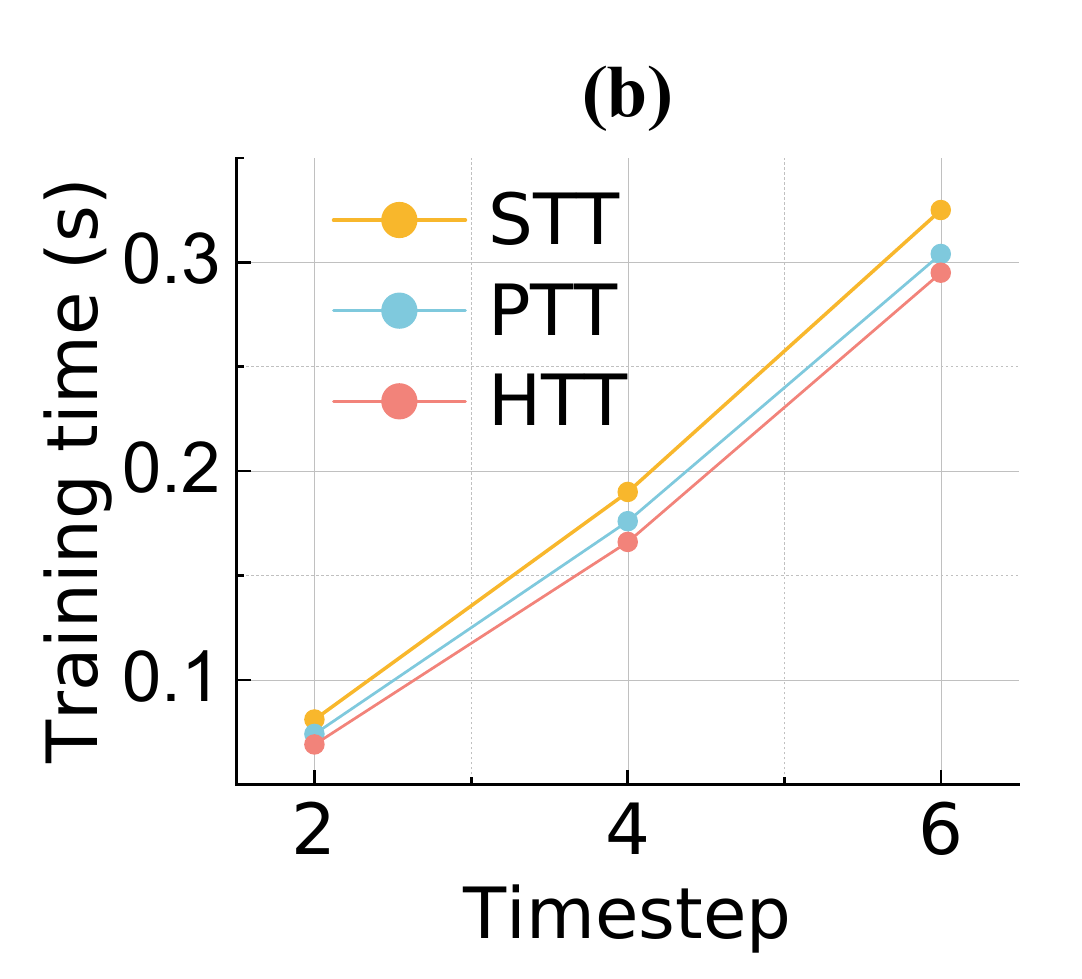}}
    \vspace{-2mm}
    \caption{Performance trends according to the timesteps. (a) Accuracy and (b) training time between STT, PTT, and HTT during the training process.}
    
    \label{fig4}
    \vspace{-0.5mm}
\end{figure}

\begin{table}[h] 
\vspace{-1.0mm}
\caption{Accuracy results based on the arrangement of full and half sub-convolutions in the HTT module.}
\vspace{-2.5mm}
\begin{center}
\begin{tabular}{c|c|c|c|c}
\hline
$t=1$ & $t=2$ & $t=3$ & $t=4$ & Accuracy (\%) \\
\hline
$F$ & $F$ & $H$ & $H$ & \textbf{91.19} \\
$H$ & $H$ & $F$ & $F$ & 90.94 \\
$H$ & $F$ & $H$ & $F$ & 90.68 \\
$F$ & $H$ & $F$ & $H$ & 90.89 \\
\hline
\end{tabular}\\
\vspace{1mm}
$F$ = full sub-convolution / $H$ = half sub-convolution
\label{tab3}
\end{center}
\vspace{-6.5mm}
\end{table}

\subsection{Ablation Study}
In this section, we perform an ablation study to gain a better understanding of the TT-SNN. Specifically, we analyze the performance of TT modules based on the timestep and the placement order of full and half sub-convolutions within the HTT modules. The ablation study is conducted with the ResNet18 and CIFAR10 dataset.

\noindent \textbf{TT Modules through Timestep: }
Given that the SNN architecture incorporates a timestep variable, it becomes crucial to examine how TT modules perform concerning this variable. Fig.~\ref{fig4} illustrates that our proposed TT modules work across different timesteps. Notably, the PTT module consistently achieves the highest accuracy, while the HTT module consistently exhibits the fastest training time across all timesteps.

\noindent \textbf{The Order of Half Sub-convolutions: }
When employing the HTT module, we put the full sub-convolutions in the early timestep and the half sub-convolutions in the later timestep. This is motivated by~\cite{kim2023exploring}, which suggests that SNN architectures tend to capture more information in the early timesteps. To further investigate this argument, we conduct experiments by altering the placement of full and half sub-convolutions within a 4-timestep ResNet18 architecture, as shown in Table \ref{tab3}. Note that we use two full sub-convolutions and two half sub-convolutions. As anticipated, when the full sub-convolutions are located in the early timestep, i.e., $t=1,2$, we achieve the highest accuracy.
\vspace{-3mm}

\section{Conclusion}
In this work, we have proposed TT-SNN architecture to gather several advantages of memory and computation costs during the SNN training. We first apply TT decomposition to SNN and modify the training pipeline to incorporate parallel computation instead of traditional sequential computation between TT-cores. Our extensive experiments on CIFAR10/100, and N-Caltech101 datasets validate the effectiveness of our efficient training technique for both static and dynamic datasets. Furthermore, our TT modules can be easily and flexibly adopted in other SNN-based convolutional architectures, enabling SNNs to maintain enhanced training efficiency and reduced computational overhead in a variety of SNN applications.
\vspace{-5.5mm}

\section*{\sc Acknowledgments}
\small 
This work was supported in part by CoCoSys, a JUMP2.0 center sponsored by DARPA and SRC, the National Science Foundation (CAREER Award, Grant \#2312366, Grant \#2318152), TII (Abu Dhabi), and the DoE MMICC center SEA-CROGS (Award \#DE-SC0023198).

\vspace{-1.5mm}
\bibliographystyle{IEEEtran}
\bibliography{main}

@article{sata,
  title={SATA: Sparsity-aware training accelerator for spiking neural networks},
  author={Yin, Ruokai and others},
  journal={TCAD},
  year={2022}
}

@article{davies2018loihi,
  title={Loihi: A neuromorphic manycore processor with on-chip learning},
  author={Davies, Mike and otherss},
  journal={IEEE Micro},
  volume={38},
  number={1},
  pages={82--99},
  year={2018},
  publisher={IEEE}
}

@article{roy2019towards,
  title={Towards spike-based machine intelligence with neuromorphic computing},
  author={Roy, Kaushik and Jaiswal, Akhilesh and Panda, Priyadarshini},
  journal={Nature},
  volume={575},
  number={7784},
  pages={607--617},
  year={2019},
  publisher={Nature Publishing Group UK London}
}

@article{hu2021advancing,
  title={Advancing Spiking Neural Networks towards Deep Residual Learning},
  author={Hu, Yifan and Deng, Lei and Wu, Yujie and Yao, Man and Li, Guoqi},
  journal={arXiv preprint arXiv:2112.08954},
  year={2021}
}

@inproceedings{diehl2015fast,
  title={Fast-classifying, high-accuracy spiking deep networks through weight and threshold balancing},
  author={Diehl, Peter U and others},
  booktitle={2015 International joint conference on neural networks (IJCNN)},
  pages={1--8},
  year={2015},
  organization={ieee}
}

@inproceedings{han2020deep,
  title={Deep spiking neural network: Energy efficiency through time based coding},
  author={Han, Bing and Roy, Kaushik},
  booktitle={European Conference on Computer Vision},
  pages={388--404},
  year={2020},
  organization={Springer}
}

@article{wu2018spatio,
  title={Spatio-temporal backpropagation for training high-performance spiking neural networks},
  author={Wu, Yujie and OTHERS},
  journal={Frontiers in neuroscience},
  year={2018},
  publisher={Frontiers Media SA}
}

@article{shrestha2018slayer,
  title={Slayer: Spike layer error reassignment in time},
  author={Shrestha, Sumit B and Orchard, Garrick},
  journal={Advances in neural information processing systems},
  volume={31},
  year={2018}
}

@inproceedings{putra2021q,
  title={Q-spinn: A framework for quantizing spiking neural networks},
  author={Putra, Rachmad Vidya Wicaksana and Shafique, Muhammad},
  booktitle={2021 International Joint Conference on Neural Networks (IJCNN)},
  pages={1--8},
  year={2021},
  organization={IEEE}
}

@inproceedings{xu2023constructing,
  title={Constructing deep spiking neural networks from artificial neural networks with knowledge distillation},
  author={Xu, Qi and others},
  booktitle={Proceedings of the IEEE/CVF Conference on Computer Vision and Pattern Recognition},
  pages={7886--7895},
  year={2023}
}

@inproceedings{kushawaha2021distilling,
  title={Distilling spikes: Knowledge distillation in spiking neural networks},
  author={Kushawaha, Ravi Kumar and Kumar, Saurabh and Banerjee, Biplab and Velmurugan, Rajbabu},
  booktitle={2020 25th International Conference on Pattern Recognition (ICPR)},
  pages={4536--4543},
  year={2021},
  organization={IEEE}
}

@article{oseledets2011tensor,
  title={Tensor-train decomposition},
  author={Oseledets, Ivan V},
  journal={SIAM Journal on Scientific Computing},
  volume={33},
  number={5},
  pages={2295--2317},
  year={2011},
  publisher={SIAM}
}

@inproceedings{ding2019acnet,
  title={Acnet: Strengthening the kernel skeletons for powerful cnn via asymmetric convolution blocks},
  author={Ding, Xiaohan and Guo, Yuchen and Ding, Guiguang and Han, Jungong},
  booktitle={Proceedings of the IEEE/CVF international conference on computer vision},
  pages={1911--1920},
  year={2019}
}

@inproceedings{zheng2021going,
  title={Going deeper with directly-trained larger spiking neural networks},
  author={Zheng, Hanle and Wu, Yujie and Deng, Lei and Hu, Yifan and Li, Guoqi},
  booktitle={Proceedings of the AAAI conference on artificial intelligence},
  volume={35},
  number={12},
  pages={11062--11070},
  year={2021}
}

@article{liu2023tensor,
  title={Tensor Decomposition for Model Reduction in Neural Networks: A Review},
  author={Liu, Xingyi and Parhi, Keshab K},
  journal={arXiv preprint arXiv:2304.13539},
  year={2023}
}

@article{cichocki2015tensor,
  title={Tensor decompositions for signal processing applications: From two-way to multiway component analysis},
  author={Cichocki, Andrzej and others},
  journal={IEEE signal processing magazine},
  volume={32},
  number={2},
  pages={145--163},
  year={2015},
  publisher={IEEE}
}

@article{wang2020compressing,
  title={Compressing 3DCNNs based on tensor train decomposition},
  author={Wang, Dingheng and Zhao, Guangshe and Li, Guoqi and Deng, Lei and Wu, Yang},
  journal={Neural Networks},
  volume={131},
  pages={215--230},
  year={2020},
  publisher={Elsevier}
}

@article{lee2021qttnet,
  title={QTTNet: Quantized tensor train neural networks for 3D object and video recognition},
  author={Lee, Donghyun and others},
  journal={Neural Networks},
  volume={141},
  pages={420--432},
  year={2021},
  publisher={Elsevier}
}

@inproceedings{gabor2022convolutional,
  title={Convolutional Neural Network Compression via Tensor-Train Decomposition on Permuted Weight Tensor with Automatic Rank Determination},
  author={Gabor, Mateusz and Zdunek, Rafa{\l}},
  booktitle={ICCS},
  pages={654--667},
  year={2022}
}

@inproceedings{kim2023exploring,
  title={Exploring temporal information dynamics in spiking neural networks},
  author={Kim, Youngeun and others},
  booktitle={Proceedings of the AAAI Conference on Artificial Intelligence},
  volume={37},
  number={7},
  pages={8308--8316},
  year={2023}
}

@article{nakajima2013global,
  title={Global analytic solution of fully-observed variational Bayesian matrix factorization},
  author={Nakajima, Shinichi and Sugiyama, Masashi and Babacan, S Derin and Tomioka, Ryota},
  journal={The Journal of Machine Learning Research},
  volume={14},
  number={1},
  pages={1--37},
  year={2013},
  publisher={JMLR. org}
}

@article{liang2021h2learn,
  title={H2learn: High-efficiency learning accelerator for high-accuracy spiking neural networks},
  author={Liang, Ling and others},
  journal={IEEE Transactions on Computer-Aided Design of Integrated Circuits and Systems},
  volume={41},
  number={11},
  pages={4782--4796},
  year={2021},
  publisher={IEEE}
}

@article{izhikevich2003simple,
  title={Simple model of spiking neurons},
  author={Izhikevich, Eugene M},
  journal={IEEE Transactions on neural networks},
  volume={14},
  number={6},
  pages={1569--1572},
  year={2003},
  publisher={IEEE}
}

@article{yin2023workload,
  title={Workload-balanced pruning for sparse spiking neural networks},
  author={Yin, Ruokai and others},
  journal={arXiv preprint arXiv:2302.06746},
  year={2023}
}

@inproceedings{li2022neuromorphic,
  title={Neuromorphic data augmentation for training spiking neural networks},
  author={Li, Yuhang and others},
  booktitle={European Conference on Computer Vision},
  pages={631--649},
  year={2022},
  organization={Springer}
}

@article{duan2022temporal,
  title={Temporal effective batch normalization in spiking neural networks},
  author={Duan, Chaoteng and Ding, Jianhao and Chen, Shiyan and Yu, Zhaofei and Huang, Tiejun},
  journal={Advances in Neural Information Processing Systems},
  volume={35},
  pages={34377--34390},
  year={2022}
}

@article{deng2022temporal,
  title={Temporal efficient training of spiking neural network via gradient re-weighting},
  author={Deng, Shikuang and Li, Yuhang and Zhang, Shanghang and Gu, Shi},
  journal={arXiv preprint arXiv:2202.11946},
  year={2022}
}

@inproceedings{narayanan2020spinalflow,
  title={SpinalFlow: An architecture and dataflow tailored for spiking neural networks},
  author={Narayanan, Surya and others},
  booktitle={2020 ACM/IEEE 47th Annual International Symposium on Computer Architecture (ISCA)},
  pages={349--362},
  year={2020},
  organization={IEEE}
}

@inproceedings{amir2017low,
  title={A low power, fully event-based gesture recognition system},
  author={Amir, Arnon and others},
  booktitle={Proceedings of the IEEE conference on computer vision and pattern recognition},
  pages={7243--7252},
  year={2017}
}

@article{orchard2015converting,
  title={Converting static image datasets to spiking neuromorphic datasets using saccades},
  author={Orchard, Garrick and Jayawant, Ajinkya and Cohen, Gregory K and Thakor, Nitish},
  journal={Frontiers in neuroscience},
  volume={9},
  pages={437},
  year={2015},
  publisher={Frontiers Media SA}
}

@inproceedings{wu2019direct,
  title={Direct training for spiking neural networks: Faster, larger, better},
  author={Wu, Yujie and others},
  booktitle={Proceedings of the AAAI conference on artificial intelligence},
  volume={33},
  number={01},
  pages={1311--1318},
  year={2019}
}

@article{yin2023mint,
  title={MINT: Multiplier-less Integer Quantization for Spiking Neural Networks},
  author={Yin, Ruokai and Li, Yuhang and Moitra, Abhishek and Panda, Priyadarshini},
  journal={arXiv preprint arXiv:2305.09850},
  year={2023}
}

\end{document}